\documentclass{article}

\usepackage{PRIMEarxiv}

\usepackage[utf8]{inputenc} % allow utf-8 input
\usepackage[T1]{fontenc}    % use 8-bit T1 fonts
\usepackage{hyperref}       % hyperlinks
\usepackage{url}            % simple URL typesetting
\usepackage{booktabs}       % professional-quality tables
\usepackage{amsfonts}       % blackboard math symbols
\usepackage{nicefrac}       % compact symbols for 1/2, etc.
\usepackage{microtype}      % microtypography
\usepackage{lipsum}
\usepackage{fancyhdr}       % header
\usepackage{graphicx}       % graphics
\graphicspath{{media/}}     % organize your images and other figures under media/ folder
\usepackage{framed}
\usepackage{multicol}
\usepackage{multirow}
\usepackage{subcaption}
\title{ALPHA: AnomaLous Physiological Health Assessment Using Large Language Models
}

\author{
  Jiankai Tang \\
  Tsinghua University \\
  Beijing\\
  \texttt{tjk19@mails.tsinghua.edu.cn} \\
   \And
  Kegang Wang \\
  Central China Normal University \\
  Wuhan\\
  \texttt{kegangwang@mails.ccnu.edu.cn} \\
   \And
   Hongming Hu \\
  Tsinghua University \\
  Beijing\\
  \texttt{huhm19@mails.tsinghua.edu.cn} \\
  \And
   Xiyuxing Zhang\\
  Tsinghua University \\
  Beijing\\
  \texttt{zxyx22@mails.tsinghua.edu.cn} \\
  \And
  Peiyu Wang \\
  Zhipu AI \\
  Beijing\\
  \texttt{peiyu.wang@aminer.cn} \\
     \And
  Xin Liu \\
 University of Washington\\
  Seattle\\
  \texttt{xliu0@cs.washington.edu} \\
     \And
  Yuntao Wang\thanks{indicates the corresponding author} \\
  Tsinghua University \\
  Beijing\\
  \texttt{yuntaowang@tsinghua.edu.cn} \\
}

\begin{document}
\maketitle

\begin{abstract}
This study concentrates on evaluating the efficacy of Large Language Models (LLMs) in healthcare, with a specific focus on their application in personal anomalous health monitoring. Our research primarily investigates the capabilities of LLMs in interpreting and analyzing physiological data obtained from FDA-approved devices. We conducted an extensive analysis using anomalous physiological data gathered in a simulated low-air-pressure plateau environment. This allowed us to assess the precision and reliability of LLMs in understanding and evaluating users' health status with notable specificity. Our findings reveal that LLMs exhibit exceptional performance in determining medical indicators, including a Mean Absolute Error (MAE) of less than 1 beat per minute for heart rate and less than 1\% for oxygen saturation (SpO2). Furthermore, the Mean Absolute Percentage Error (MAPE) for these evaluations remained below 1\%, with the overall accuracy of health assessments surpassing 85\%. In image analysis tasks, such as interpreting photoplethysmography (PPG) data, our specially adapted GPT models demonstrated remarkable proficiency, achieving less than 1 bpm error in cycle count and  7.28 MAE for heart rate estimation. This study highlights LLMs' dual role as health data analysis tools and pivotal elements in advanced AI health assistants, offering personalized health insights and recommendations within the future health assistant framework as Figure ~\ref{fig: framework}.

\end{abstract}

% keywords can be removed

\section{Introduction}
Large language models (LLMs), exemplified by systems such as GPT and GLM, have exhibited remarkable proficiency in capturing and generating diverse knowledge, attributed to their scaled neural architectures characterized by a proliferation of parameters and extensive training data~(e.g., \cite{brown2020language,chowdhery2022palm,openai2023gpt4,cluade2023cluade,du2022glm,zeng2022glm}). These models have found applications across domains including software engineering, content generation~(e.g., \cite{chen2021evaluating,chung2022talebrush}), and healthcare~(e.g., \cite{singhal2022large}). Nevertheless, while medical LLMs have demonstrated competence in comprehending intricate medical knowledge~(e.g., \cite{gu2021domain,singhal2022large}), their potential for analyzing physiological and behavioral time series data, crucial in consumer health applications, remains a realm yet to be fully explored.

Utilizing Large Language Models (LLMs) in healthcare presents both intriguing possibilities and inherent challenges. While LLMs demonstrate proficiency in question-answering tasks, their capability in medical diagnostics can be constrained by their limited foundation in physiological principles, as pointed out by Lubitz et al. \cite{lubitz2022detection}. However, when combined with physiological data, these models can potentially offer richer medical insights, as highlighted by studies from Ferguson et al.\cite{ferguson2022effectiveness} and Liu et al. \cite{liu2023large}. Notably, LLMs are capable of extracting valuable information from even sparse numerical health datasets, which showcases their potential in augmenting health-related decision-making\cite{montagna2023data}. Nevertheless, a deeper investigation is required to confirm their effectiveness in analyzing physiological and behavioral data, especially multimodal data\cite{mesko2023impact}.

%Addressing the healthcare domain, particularly in clinical diagnostics and consumer electronics, demands rigorous physiological monitoring. Extracting meaningful health insights from wearable devices and cost-effective sensing platforms has gained prominence. Amidst these prospects, the application of LLMs introduces a distinctive set of challenges. These encompass the need to surmount the inherent illusions surrounding these models and ensure their mathematical rigor, which in turn engenders a cautious stance among researchers towards their healthcare application.

This paper seeks to explore the potential of LLMs in understanding physiological vital signs and concisely representing users' health statuses, contextualized within everyday scenarios. Specifically, we evaluate the mathematical precision and medical diagnostic capabilities of LLMs, leveraging authentic clinical data and synthetically generated scenarios, thereby establishing the feasibility of LLMs as proficient health assistants. Results show that the LLMs could achieve below 1 MAE in the vitals calculation, over 85\% accuracy in health status assessment, and below 1 MAE in the visual cycle counting.

%In the following research, we will address the challenges inherent in incorporating LLMs into health assistant applications. These challenges encompass the fusion of diverse data modalities, reliable dissemination of health advisories, and the ethical implications of utilizing generative AI in healthcare. Mitigating these challenges is pivotal for the responsible integration of LLMs in healthcare contexts.
\begin{figure}
    \centering
  \includegraphics[width=\textwidth]{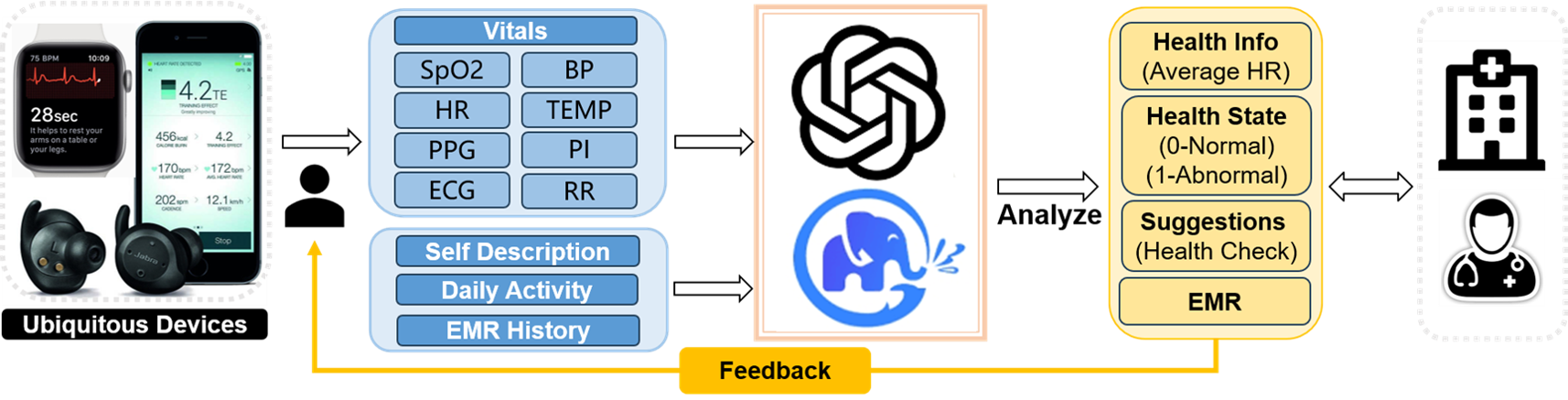}

  % \vspace{-8pt}
  \caption{{Framework of HealthCarer}}
    \label{fig: framework}
  % \Description{The pipeline of utilizing vitals in health care with Cluade LLM.}
  % \vspace{-8pt}
  {\small{Interpretation of medical terms: SpO2: Oxygen Saturation, BP: Blood Pressure, HR: Heart Rate, TEMP: Body Temperature, PPG: Photoplethysmogram, PI: Perfusion Index, ECG: Electrocardiogram, RR: Respiration Rate, EMR: Electronic Medical Record}}
\end{figure}

% \section{Motivation}
% % \label{sec:headings}
% In recent years, popular wearable devices like smartwatches and earphones have gained momentum, connecting seamlessly to smartphones via Bluetooth. Many of these devices can measure physiological factors such as PPG, ECG and heart rate\cite{Boukhayma_2021}, enabling users to monitor their well-being in real time.

% Simultaneously, mobile devices, equipped with Large Language Models (LLMs), gather user preferences and daily activity, merging this data with medical records. This integrated dataset fuels the LLM's medical analysis, offering insights, defining well-being, suggesting healthy habits, and providing motivation. These insights are stored in electronic health records, fostering a connection with medical systems, allowing feedback and advice from the medical community for a comprehensive healthcare approach. As illustrated in Figure~\ref{fig: framework}, our system framework could validate our approach.
% \lipsum[4] See Section \ref{sec:headings}.

\section{Challenges}
This research delves into the complex challenges of integrating Large Language Models (LLMs) in health assistant applications, a critical step for the advancement of medical technology. These challenges range from effectively combining various sensor modalities to ensuring the reliable dissemination of health advisories and grappling with the ethical implications of using generative AI in healthcare settings. The careful mitigation of these issues is essential for the responsible and effective integration of LLMs, paving the way for a new era in healthcare technology.

\subsection{Modality Fusion}
Integrating multimodal models like Langchain\cite{Chase2022langchain}, BLIP-2\cite{li2023blip2}, and LLaVA\cite{liu2023llava} into health assistants presents a formidable challenge. These systems must be enabled to process and represent data from multiple modalities, such as images and audio, as well as seamlessly convert them into coherent, actionable medical insights. The key lies in achieving precision in data representation and maintaining efficiency, which are crucial for their effective application in healthcare settings. This integration demands a nuanced understanding of both technology and healthcare needs.

\subsection{Responsible Generative AI}
The use of LLMs in healthcare introduces significant challenges, particularly due to issues like 'hallucinations'\cite{Hallucination_llm}. Striking a balance between harnessing the capabilities of generative AI and ensuring the accuracy and reliability of the information it provides is paramount, especially in sensitive areas such as patient recovery and medical record management. It's essential to establish robust mechanisms to validate AI-generated advice and maintain transparency in AI-driven decision-making processes.

\subsection{Patient Privacy}
Ensuring patient privacy in the integration of generative AI into health assistants is a critical challenge\cite{jo2023understanding}. The use of LLMs for processing sensitive information, including physiological signals and medical records, heightens concerns regarding data security and confidentiality. This challenge involves not only protecting against unauthorized data breaches but also ensuring compliance with stringent regulations like HIPAA\footnote{https://www.cdc.gov/phlp/publications/topic/hipaa.html}. The goal is to balance the transformative benefits of AI in healthcare with the imperative of safeguarding patient confidentiality, necessitating the development of advanced encryption methods, strict access controls, and clear data usage policies.

\section{Experiments}
Tasks are chosen to validate the performance of mainstream LLMs in health-related tasks:(1) calculate average vital information from raw biomedical sensor signals(e.g., the average heart rate for 60 seconds) (2) assess health statuses based on medical knowledge. To do the experiments, open-source physiological datasets(e.g., \cite{tang2023mmpd,wang2023physbench} )are first tested in our system. However, these datasets do not include data from individuals with abnormal health conditions (e.g., arrhythmia), which makes it hard to tell whether the LLMs output the correct assessment. To fill the gap, we conduct a user study under low air pressure to simulate the environment of a highland and collect multiple physiological data for 12 subject tests. Our experiment was approved by the Institutional Review Board (No.20230076). In the low-air condition, blood oxygen may decrease even to hypoxia syndrome, and heart rate, and respiration rate would increase, which enables the collection of physiological data in instances of abnormal health conditions. The detailed results are shown in the following. The related code is publicly available in our GitHub repository: \href{https://github.com/McJackTang/LLM-HealthAssistant}{https://github.com/McJackTang/LLM-HealthAssistant}.

\subsection{Vitals Calculation}
\label{sec: vitals_calculation}
In the pursuit of vital calculation, the primary objective involves tasking the large language model with calculating the average values of vital signs (e.g., HR, SpO2) over continuous time intervals. This experimental study encompasses two distinct tasks: (1) Single-Task, where a solitary vital sign is provided as input along with a corresponding prompt to compute a singular value; (2) Multi-Task, which involves inputting two vital signs simultaneously and prompting the model to compute two values concurrently. The Single-Task is specifically designed to evaluate the accuracy and reliability of the Large Language Model (LLM), while the Multi-Task is intended to assess the model's proficiency in processing and interpreting complex information efficiently. Our findings indicate that LLMs demonstrate commendable performance in both of these tasks. As depicted in Table~\ref{table:vitals_calcu_results}, the least Mean Absolute Error (MAE) and Mean Absolute Percentage Error (MAPE) metrics remain consistently below 1 for both input formats. Notably, GPT outperforms GLM in the majority of vital calculation tasks, with the exception being the single SpO2 calculation task spanning 120 seconds. This discrepancy suggests that GPT exhibits superior floating-point capabilities, whereas GLM tends to produce integer outputs.
\begin{table}[h]
\centering
\small

\caption{Vitals Calculation On Average Vitals}
\label{table:vitals_calcu_results}

% \tiny % 使用紧凑字体
\begin{tabular}{c|c|cc|cc|cc|cc} % 减少列宽度
\toprule
\multirow{2}*{\textbf{Model}} & \multirow{2}*{\textbf{Period(s)}} & \multicolumn{2}{c}{\textbf{Single HR}} & \multicolumn{2}{c}{\textbf{Single SpO2}} & \multicolumn{2}{c}{\textbf{Multi HR}} & \multicolumn{2}{c}{\textbf{Multi SpO2}} 
 \\
% \cmidrule(lr){2-3} \cmidrule(lr){4-5} \cmidrule(lr){6-9} % 使用更轻的分割线
&& MAE$\downarrow$ & MAPE$\downarrow$ & MAE$\downarrow$ & MAPE$\downarrow$ & MAE$\downarrow$ & MAPE$\downarrow$ & MAE$\downarrow$ & MAPE$\downarrow$ \\
\midrule
GLM&30 & 1.18 & 1.62 & 0.52 & 0.56 & 1.73 & 2.39 & 0.73 & 0.78 \\
GPT&30 & 0.48 & 0.66 & 0.36 & 0.38 & 0.72 & 1.01 & 0.65 & 0.69 \\
GLM&60 & 1.24 & 1.73 & 0.52 & 0.55 & 1.72 & 2.39 & 0.66 & 0.70 \\
GPT&60 & \textbf{0.44} & \textbf{0.61} & 0.33 & 0.35 & 0.58 & 0.82 & 0.53 & 0.56 \\
GLM&120 & 1.26 & 1.78 & \textbf{0.21} & \textbf{0.22 }& 2.22 & 3.09 & 0.33 & 0.34 \\
GPT&120 & 0.56 & 0.79 & 0.24 & 0.25 & 0.46 & 0.65 & \textbf{0.27} & \textbf{0.28} \\
% 60 & 0.29 & 0.40 & 0.25 & 0.26 & 0.58 & 0.82 & 0.79 & 0.84 \\
\bottomrule
\end{tabular}
\\
\footnotesize{GLM stands for GLM-2-Pro and GPT stands for GPT-3.5-Turbo at October 2023. MAE = Mean Absolute Error in HR/SpO2 estimation (HR:Beats/Min,SpO2 :\%), MAPE = Mean Absolute Percentage Error in HR/SpO2 estimation (\%).}
\end{table}

\subsection{Health Status Assessment}
The objective of the multi-classification mission is to evaluate the health status based on the analysis of vital signs, encompassing various distinct and blended vital sign datasets in a manner similar to the structure outlined in Section~\ref{sec: vitals_calculation}. This assessment adheres to the established guidelines\footnote{https://www.who.int/news-room/questions-and-answers/item/oxygen} provided by the World Health Organization (WHO). For example, an abnormal heart rate (HR) exceeding 100 and an extremely abnormal state surpassing 130 are indicative of deviations from the norm. Similarly, deviations in oxygen saturation (SpO2) are considered abnormal when falling below 95 and extremely abnormal when dropping below 92. When dealing with amalgamated health indicators, the designation of normal health for the multi-task scenario is contingent upon both HR and SpO2 readings concurring as normal. To be noticed, abnormal health statuses in SpO2 are more prevalent than those in heart rate in the distribution of our dataset. As illustrated in Table~\ref{table:health_status_results}, GLM demonstrates superior performance in the assessment of the single heart rate task, while GPT performs better in the single SpO2 task exhibiting its adaptability to changes.  It is noteworthy that GLM appears to be capable of handling multimodal data, as evidenced by the increase in accuracy when heart rate and SpO2 data are processed concurrently.
\begin{table}[h]
\centering
\small
\caption{Accuracy of Health Status Classification}
\label{table:health_status_results}
% \tiny % 使用紧凑字体
\begin{tabular}{c|c|c|c|c} % 减少列宽度
\toprule
\multirow{1}*{\textbf{Model}}&\multirow{1}*{\textbf{Period(s)}} & \multicolumn{1}{c}{\textbf{Single HR Health Acc$\uparrow$ }} & \multicolumn{1}{c}{\textbf{Single SpO2 Health Acc$\uparrow$}} & \multicolumn{1}{c}{\textbf{Multi Health Acc$\uparrow$}} 
 \\
 % 使用更轻的分割线

\midrule
GLM&30   & 0.78 & 0.72 & \textbf{0.87} \\
GPT&30   & 0.68 & \textbf{0.81} & 0.76 \\
GLM&60 &  0.75 & 0.75 & 0.83 \\
GPT&60 &  0.75 & \textbf{0.81} & 0.75 \\
GLM&120 &  \textbf{0.87} & 0.67 & 0.80 \\
GPT&120 & 0.73 & 0.80 & \textbf{0.87} \\
% 60 & 0.97 & 0.61 & 0.83  \\
\bottomrule
\end{tabular}
\\
\footnotesize{GLM stands for GLM-2-Pro and GPT stands for GPT-3.5-Turbo at October 2023. Accuracy(Acc) calculates the ratio of correct predictions to total predictions, within a range of 0 to 1.}
\end{table}

\subsection{Medical Image Analysis}

\begin{figure}  
    \centering  
    % \begin{subfigure}{0.32\linewidth}  
    % \centering
    %     \includegraphics[width=0.8\linewidth]{Visual Counter.png}  
    %     \caption{Health assessment with vitals from wearables}  
    %     \label{fig:sub1}  
    % \end{subfigure}  
    \begin{subfigure}{0.45\linewidth}  
    \centering
        \includegraphics[width=1.0\linewidth]{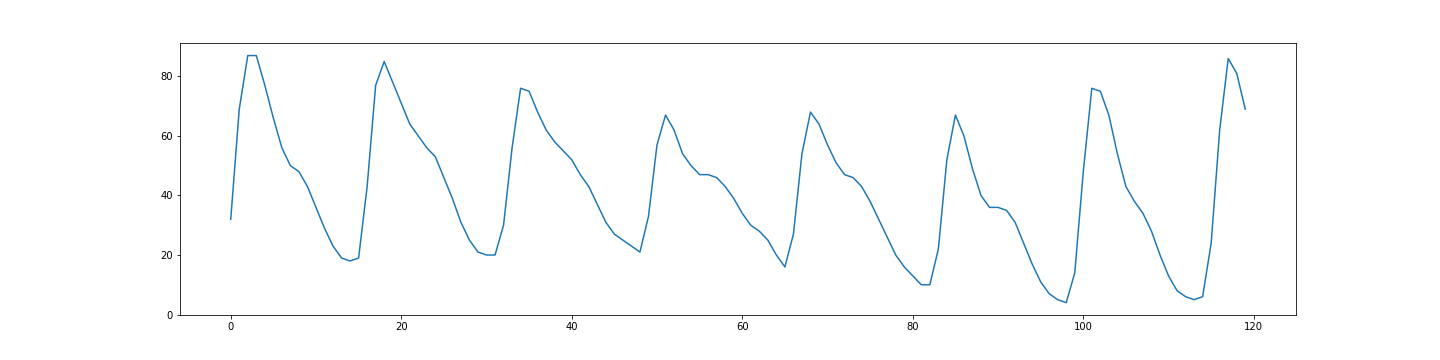}  
        \caption{Sample of bvp}  
        \label{fig:sub1}  
        % 创建一个带边框的文本框

    \begin{framed}
Here is a graph of the PPG wave for 10 seconds. Please count the number of peaks and troughs, and calculate the heart rate.
\end{framed}

    \end{subfigure} 
    \begin{subfigure}{0.45\linewidth} 
    \centering
        \includegraphics[width=0.8\linewidth]{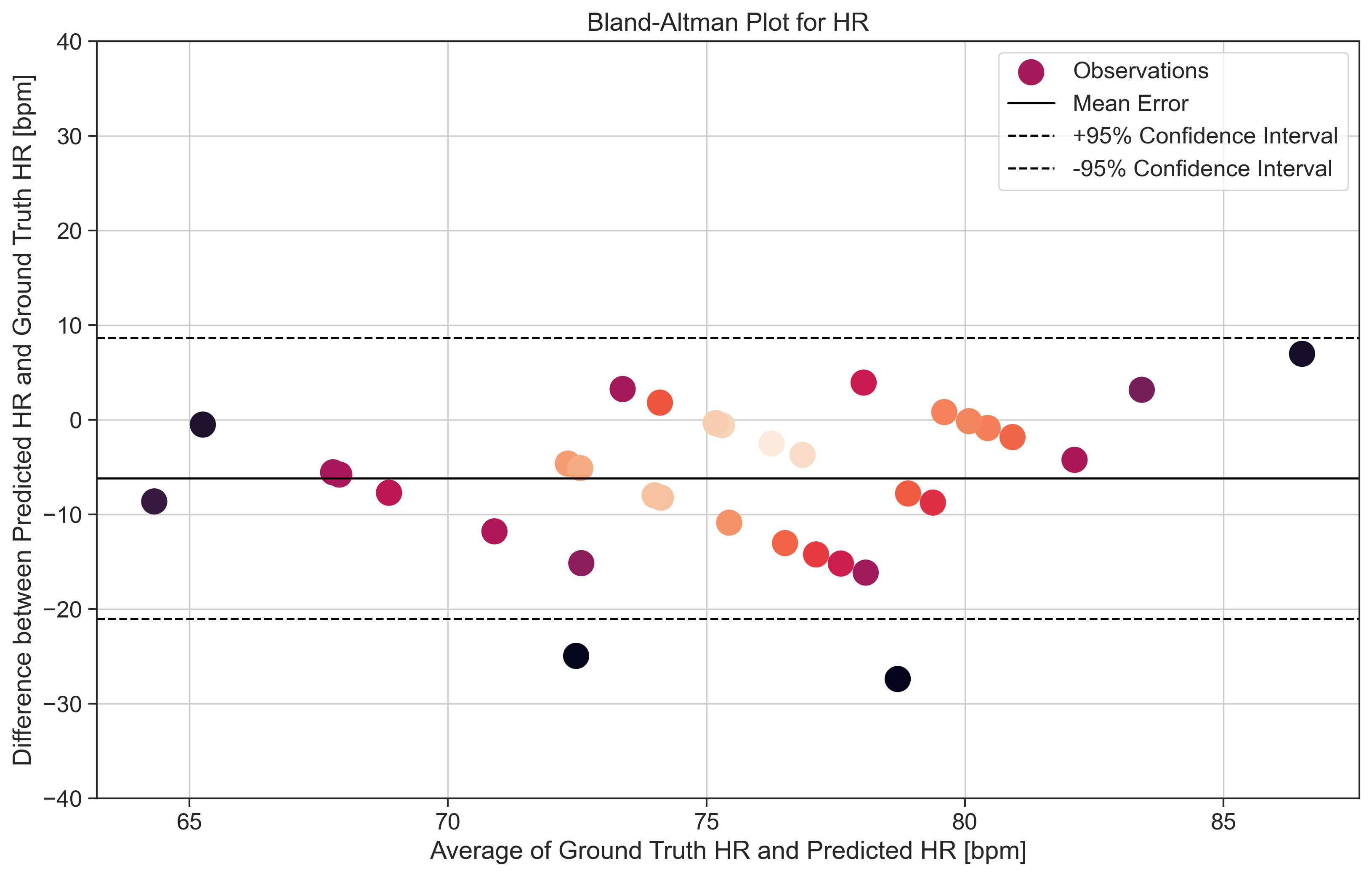}  
        \caption{Bland\_Altman Plot for heart rate estimation}  
        \label{fig:sub2}  
    \end{subfigure}  
    \caption{Visual Counter Sample and Result}  
    \label{fig:Visual Counter}  
\end{figure}

In addition to textual information, dynamic medical vital images such as PPG or ECG play a crucial role in assessing the health condition of people. To evaluate the ability of the Large Language Model (LLM) to comprehend PPG graphs, a specialized GPT model was tasked with counting the cycles in the images and subsequently calculating the heart rate through analysis. To the best of our knowledge, we are the first to design and fine-tune a GPTs model referred to as the "visual counter" accessible at (\href{https://chat.openai.com/g/g-SR8nCXyWI-visual-counter}{https://chat.openai.com/g/g-SR8nCXyWI-visual-counter}).  Our tailored GPTs model could achieve 0.556 MAE and 0.889 MAE for peak and trough count respectively, which leads to 7.283 MAE and 9\% MAPE for heart rate estimation, as illustrated in  Figure ~\ref{fig:Visual Counter}.

% \begin{figure}
%     \centering
%   \includegraphics[width=\textwidth]{hr_bland_altman.png}
%   \label{fig: visual_count}
%   % \vspace{-8pt}
%   \caption{{Bland\_Altman Plot for heart rate estimation from visual count}}

% \end{figure}

\section{Discussion}
\subsection{AI Agents}
The inherent randomness and perceived unpredictability of large language models have traditionally posed challenges when it comes to performing complex calculations. Nevertheless, recent advancements in AI agents have emerged, making it increasingly feasible for these models to handle intricate numerical tasks. When the Large Language Model (LLM) is tasked solely with understanding user intentions and is empowered to command the execution of executable programs, it becomes evident that it can achieve a higher degree of accuracy in numerical calculations.
\subsection{Applications}
\begin{figure}  
    \centering  
    \begin{subfigure}{0.45\linewidth}  
    \centering
        \includegraphics[width=0.8\linewidth]{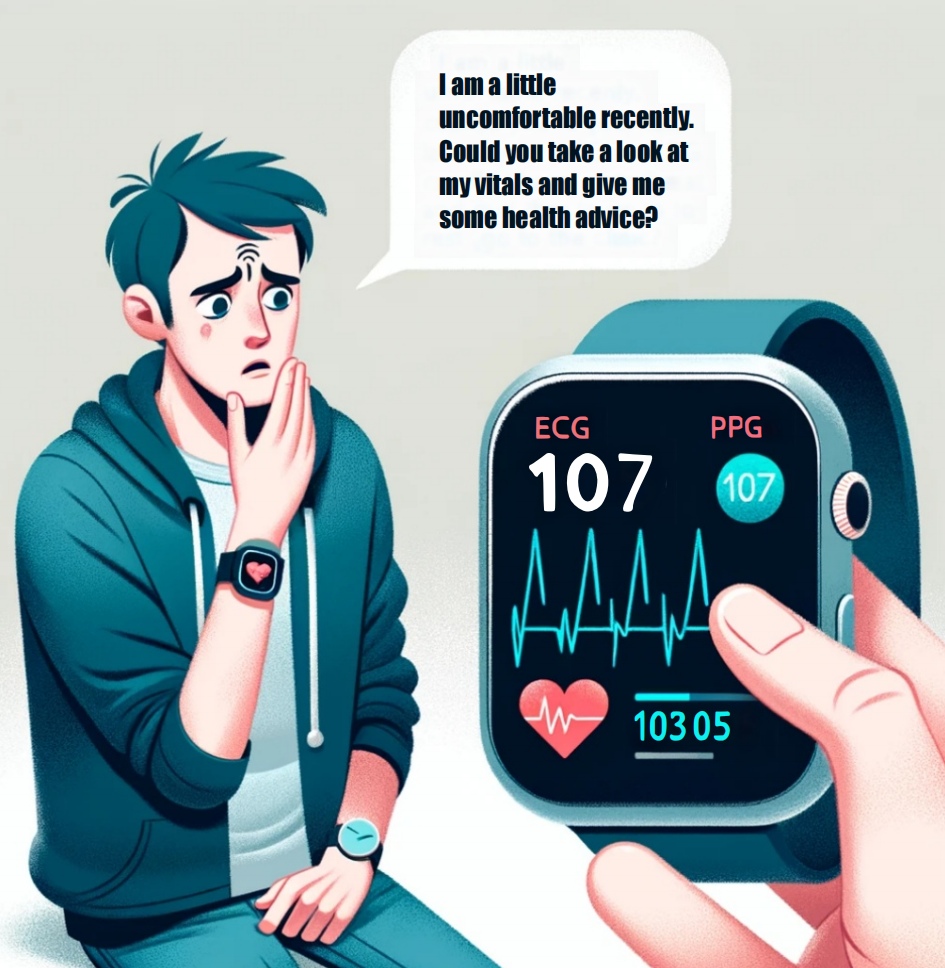}  
        \caption{Health assessment with vitals from wearables}  
        \label{fig:sub1}  
    \end{subfigure}  
    \begin{subfigure}{0.45\linewidth} 
    \centering
        \includegraphics[width=0.8\linewidth]{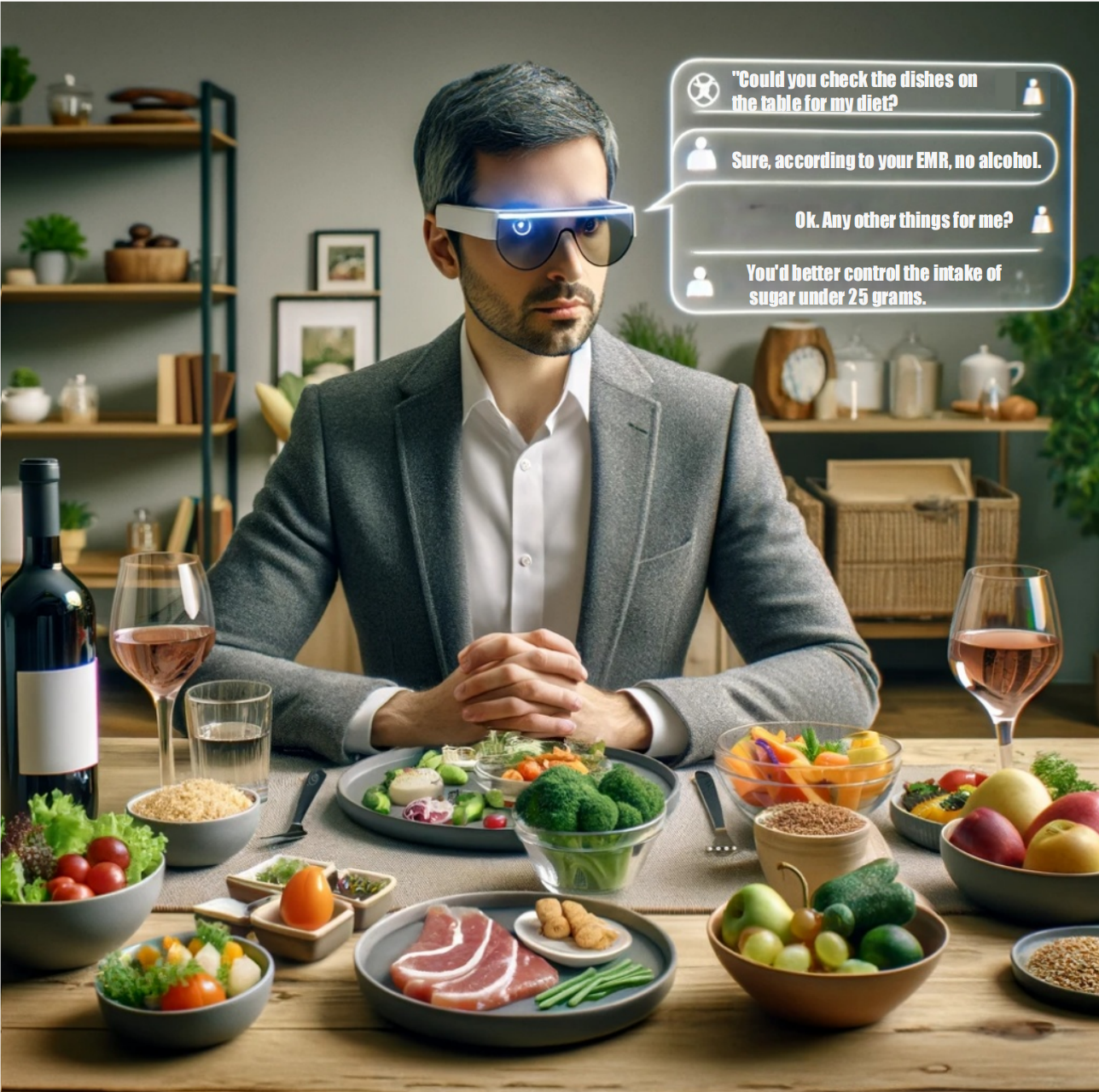}  
        \caption{Diet reminders using visual information}  
        \label{fig:sub2}  
    \end{subfigure}  
    \caption{Applications on daily life} 
    \label{fig:applications}  
\end{figure}

Forwardly, the Large Language Model (LLM) goes beyond being a mere diagnostic tool, as it has the potential to offer valuable advice and care to users. When integrated with wearable devices, the LLM has the capability to function as your personal health manager, connecting with the real world. Here, we present several scenarios to illustrate how the LLM could assist us in various everyday applications, including providing health suggestions and offering reminders for recovery, as shown in Figure ~\ref{fig:applications}\footnote{Generated with GPT Plus and Modified} .
\begin{center}
\begin{minipage}{0.48\textwidth}
\begin{framed}
\centering
\textit{User: "I am a little uncomfortable recently. Could you take a look at my vitals and give me some health advice?"}
\end{framed}
\end{minipage}
\hfill
\begin{minipage}{0.48\textwidth}
\begin{framed}
\centering
\textit{LLM: "Sure, according to your ECG and PPG, you may have atrial fibrillation. It would be better for you to take more rest and go to the clinic."}
\end{framed}
\end{minipage}
\end{center}

\begin{center}
\begin{minipage}{0.48\textwidth}
\begin{framed}
\centering
\textit{User: "Could you check the dishes on the table, according to the doctor's suggestions, which ones can't I eat?"}
\end{framed}
\end{minipage}
\hfill
\begin{minipage}{0.48\textwidth}
\begin{framed}
\centering
\textit{LLM: "According to your EMR (diabetes), there is no harm among this except for alcohol. But you'd better control the intake of sugar under 25 grams."}
\end{framed}
\end{minipage}
\end{center}

\subsection{Future Work}

In addition to performing mathematical calculations using raw text input with the Large Language Model (LLM), another promising avenue for the future lies in the development of scheduling agents capable of generating executable code. When utilizing LLM agents for such purposes, the pivotal factors for successful execution will include intricate task decomposition and rigorous stage verification. Meanwhile, in the realm of medical applications, the ability to comprehend and interpret medical graphs may hold even greater significance, particularly when it comes to aligning and integrating diverse medical information across various devices and sources.

\section{Conclusions}

In this study, we evaluate the capabilities of large language models to deal with anomalous physiological data with three unique tasks. Our contributions encompass the following key aspects: (1) Opensource Data and Code for Abnormal Health Condition Assessment: We provide an open-source dataset and code resources to facilitate the assessment of abnormal health conditions, promoting transparency and collaboration in the field. (2) Validation of Vitals Calculation using LLM: We present the proficiency of large language models (LLMs) in the precise assessment of vital signs, thereby delineating their substantial promise for deployment in healthcare settings.  (3) Health Status Analysis with Vital Signs by LLMs: We validate the LLMs' capability to analyze abnormal health status based on vital sign data, showcasing its potential as a diagnostic tool. (4) Experiments on Medical Image Information Recognition: We are the first to conduct experiments to assess the LLM's ability to extract valuable information from medical images with GPTs, expanding its utility in the medical field.

\section*{Acknowledges}
% This work is advised by Yuntao Wang and Xin Liu. Special thanks to Xin Liu and Yuntao Wang for their guidance and help.

This work is supported by the Natural Science Foundation of China (NSFC) under Grant No. 62132010 and No. 62002198, Young Elite Scientists Sponsorship Program by CAST under Grant No.2021QNRC001, Tsinghua University Initiative Scientific Research Program, Beijing Natural Science Foundation, Beijing Key Lab of Networked Multimedia, and Institute for Artificial Intelligence, Tsinghua University.

%Bibliography
\bibliographystyle{unsrt}  
\bibliography{references}

\begin{thebibliography}{10}

\bibitem{brown2020language}
Tom Brown, Benjamin Mann, Nick Ryder, Melanie Subbiah, Jared~D Kaplan, Prafulla
  Dhariwal, Arvind Neelakantan, Pranav Shyam, Girish Sastry, Amanda Askell,
  et~al.
\newblock Language models are few-shot learners.
\newblock {\em Advances in neural information processing systems},
  33:1877--1901, 2020.

\bibitem{chowdhery2022palm}
Aakanksha Chowdhery, Sharan Narang, Jacob Devlin, Maarten Bosma, Gaurav Mishra,
  Adam Roberts, Paul Barham, Hyung~Won Chung, Charles Sutton, Sebastian
  Gehrmann, et~al.
\newblock {PaLM}: Scaling language modeling with pathways.
\newblock {\em arXiv preprint arXiv:2204.02311}, 2022.

\bibitem{openai2023gpt4}
OpenAI.
\newblock Gpt-4 technical report, 2023.

\bibitem{cluade2023cluade}
{Claude}.
\newblock {Claude}.
\newblock \url{www.anthropic.com}, 2023.
\newblock Conversational AI assistant.

\bibitem{du2022glm}
Zhengxiao Du, Yujie Qian, Xiao Liu, Ming Ding, Jiezhong Qiu, Zhilin Yang, and
  Jie Tang.
\newblock Glm: General language model pretraining with autoregressive blank
  infilling.
\newblock In {\em Proceedings of the 60th Annual Meeting of the Association for
  Computational Linguistics (Volume 1: Long Papers)}, pages 320--335, 2022.

\bibitem{zeng2022glm}
Aohan Zeng, Xiao Liu, Zhengxiao Du, Zihan Wang, Hanyu Lai, Ming Ding, Zhuoyi
  Yang, Yifan Xu, Wendi Zheng, Xiao Xia, et~al.
\newblock Glm-130b: An open bilingual pre-trained model.
\newblock {\em arXiv preprint arXiv:2210.02414}, 2022.

\bibitem{chen2021evaluating}
Mark Chen, Jerry Tworek, Heewoo Jun, Qiming Yuan, Henrique Ponde de~Oliveira
  Pinto, Jared Kaplan, Harri Edwards, Yuri Burda, Nicholas Joseph, Greg
  Brockman, et~al.
\newblock Evaluating large language models trained on code.
\newblock {\em arXiv preprint arXiv:2107.03374}, 2021.

\bibitem{chung2022talebrush}
John Joon~Young Chung, Wooseok Kim, Kang~Min Yoo, Hwaran Lee, Eytan Adar, and
  Minsuk Chang.
\newblock Talebrush: sketching stories with generative pretrained language
  models.
\newblock In {\em Proceedings of the 2022 CHI Conference on Human Factors in
  Computing Systems}, pages 1--19, 2022.

\bibitem{singhal2022large}
Karan Singhal, Shekoofeh Azizi, Tao Tu, S~Sara Mahdavi, Jason Wei, Hyung~Won
  Chung, Nathan Scales, Ajay Tanwani, Heather Cole-Lewis, Stephen Pfohl, et~al.
\newblock Large language models encode clinical knowledge.
\newblock {\em arXiv preprint arXiv:2212.13138}, 2022.

\bibitem{gu2021domain}
Yu~Gu, Robert Tinn, Hao Cheng, Michael Lucas, Naoto Usuyama, Xiaodong Liu,
  Tristan Naumann, Jianfeng Gao, and Hoifung Poon.
\newblock Domain-specific language model pretraining for biomedical natural
  language processing.
\newblock {\em ACM Transactions on Computing for Healthcare (HEALTH)},
  3(1):1--23, 2021.

\bibitem{lubitz2022detection}
Steven~A Lubitz, Anthony~Z Faranesh, Caitlin Selvaggi, Steven~J Atlas, David~D
  McManus, Daniel~E Singer, Sherry Pagoto, Michael~V McConnell, Alexandros
  Pantelopoulos, and Andrea~S Foulkes.
\newblock Detection of atrial fibrillation in a large population using wearable
  devices: the fitbit heart study.
\newblock {\em Circulation}, 146(19):1415--1424, 2022.

\bibitem{ferguson2022effectiveness}
Ty~Ferguson, Timothy Olds, Rachel Curtis, Henry Blake, Alyson~J Crozier, Kylie
  Dankiw, Dorothea Dumuid, Daiki Kasai, Edward O'Connor, Rosa Virgara, et~al.
\newblock Effectiveness of wearable activity trackers to increase physical
  activity and improve health: a systematic review of systematic reviews and
  meta-analyses.
\newblock {\em The Lancet Digital Health}, 4(8):e615--e626, 2022.

\bibitem{liu2023large}
Xin Liu, Daniel McDuff, Geza Kovacs, Isaac Galatzer-Levy, Jacob Sunshine,
  Jiening Zhan, Ming-Zher Poh, Shun Liao, Paolo~Di Achille, and Shwetak Patel.
\newblock Large language models are few-shot health learners, 2023.

\bibitem{montagna2023data}
Sara Montagna, Stefano Ferretti, Lorenz~Cuno Klopfenstein, Antonio Florio, and
  Martino~Francesco Pengo.
\newblock Data decentralisation of llm-based chatbot systems in chronic disease
  self-management.
\newblock In {\em Proceedings of the 2023 ACM Conference on Information
  Technology for Social Good}, pages 205--212, 2023.

\bibitem{mesko2023impact}
Bertalan Mesk{\'o}.
\newblock The impact of multimodal large language models on health care’s
  future.
\newblock {\em Journal of Medical Internet Research}, 25:e52865, 2023.

\bibitem{Chase2022langchain}
Harrison Chase.
\newblock {LangChain}, 2022.
\newblock LangChain AI.

\bibitem{li2023blip2}
Junnan Li, Dongxu Li, Silvio Savarese, and Steven Hoi.
\newblock Blip-2: Bootstrapping language-image pre-training with frozen image
  encoders and large language models, 2023.

\bibitem{liu2023llava}
Haotian Liu, Chunyuan Li, Qingyang Wu, and Yong~Jae Lee.
\newblock Visual instruction tuning.
\newblock 2023.

\bibitem{Hallucination_llm}
Ziwei Ji, Nayeon Lee, Rita Frieske, Tiezheng Yu, Dan Su, Yan Xu, Etsuko Ishii,
  Ye~Jin Bang, Andrea Madotto, and Pascale Fung.
\newblock Survey of hallucination in natural language generation.
\newblock {\em {ACM} Computing Surveys}, 55(12):1--38, mar 2023.

\bibitem{jo2023understanding}
Eunkyung Jo, Daniel~A Epstein, Hyunhoon Jung, and Young-Ho Kim.
\newblock Understanding the benefits and challenges of deploying conversational
  ai leveraging large language models for public health intervention.
\newblock In {\em Proceedings of the 2023 CHI Conference on Human Factors in
  Computing Systems}, pages 1--16, 2023.

\bibitem{tang2023mmpd}
Jiankai Tang, Kequan Chen, Yuntao Wang, Yuanchun Shi, Shwetak Patel, Daniel
  McDuff, and Xin Liu.
\newblock Mmpd: Multi-domain mobile video physiology dataset, 2023.

\bibitem{wang2023physbench}
Kegang Wang, Yantao Wei, Mingwen Tong, Jie Gao, Yi~Tian, YuJian Ma, and
  ZhongJin Zhao.
\newblock Physbench: A benchmark framework for remote physiological sensing
  with new dataset and baseline, 2023.

\end{thebibliography}

\end{document}